\documentclass[10pt,twocolumn,letterpaper]{article}

\usepackage{cvpr}
\usepackage{times}
\usepackage{epsfig}
\usepackage{graphicx}
\usepackage{amsmath}
\usepackage{amssymb}
\usepackage{multirow}
\usepackage{hhline}

\usepackage{subcaption}
\usepackage{float}
\usepackage{tabularx}

\usepackage{soul}
\newcolumntype{Y}{>{\centering\arraybackslash}X}

\makeatletter
\def\andd{%
  \end{tabular}%
  \\
  \begin{tabular}[t]{c}}
\makeatother

\usepackage[pagebackref=true,breaklinks=true,letterpaper=true,colorlinks,bookmarks=false]{hyperref}

\cvprfinalcopy 

\def\cvprPaperID{3392} 

\newcommand{\comment}[1]{\ignorespaces}

\ifcvprfinal\fi
\begin{document}

\title{In Teacher We Trust:\\Learning Compressed Models for Pedestrian Detection}

\author{Jonathan Shen $^1$ \\
Carnegie Mellon University\\
{\tt\small jshen2@andrew.cmu.edu}\\
\and
Noranart Vesdapunt $^1$\\
Carnegie Mellon University\\
{\tt\small nvesdapu@andrew.cmu.edu}\\
\andd
Vishnu N.~Boddeti\\
Michigan State University\\
{\tt\small vishnu@msu.edu}\\
\and
Kris M. Kitani\\
Carnegie Mellon University\\
{\tt\small kkitani@cs.cmu.edu}\\
}

\maketitle

\footnotetext[1]{Contributed equally.}


\begin{abstract}
Deep convolutional neural networks continue to advance the state-of-the-art in many domains as they grow bigger and more complex. It has been observed that many of the parameters of a large network are redundant, allowing for the possibility of learning a smaller network that mimics the outputs of the large network through a process called Knowledge Distillation. We show, however, that standard Knowledge Distillation is not effective for learning small models for the task of pedestrian detection. To improve this process, we introduce a higher-dimensional hint layer to increase information flow. We also estimate the variance in the outputs of the large network and propose a loss function to incorporate this uncertainty. Finally, we attempt to boost the complexity of the small network without increasing its size by using as input hand-designed features that have been demonstrated to be effective for pedestrian detection. We succeed in training a model that contains $400\times$ fewer parameters than the large network while outperforming AlexNet on the Caltech Pedestrian Dataset.

\end{abstract}

\begin{figure}[t]
    \centering
    \begin{subfigure}[b]{\linewidth}  
        \centering 
        \includegraphics[width=\linewidth]{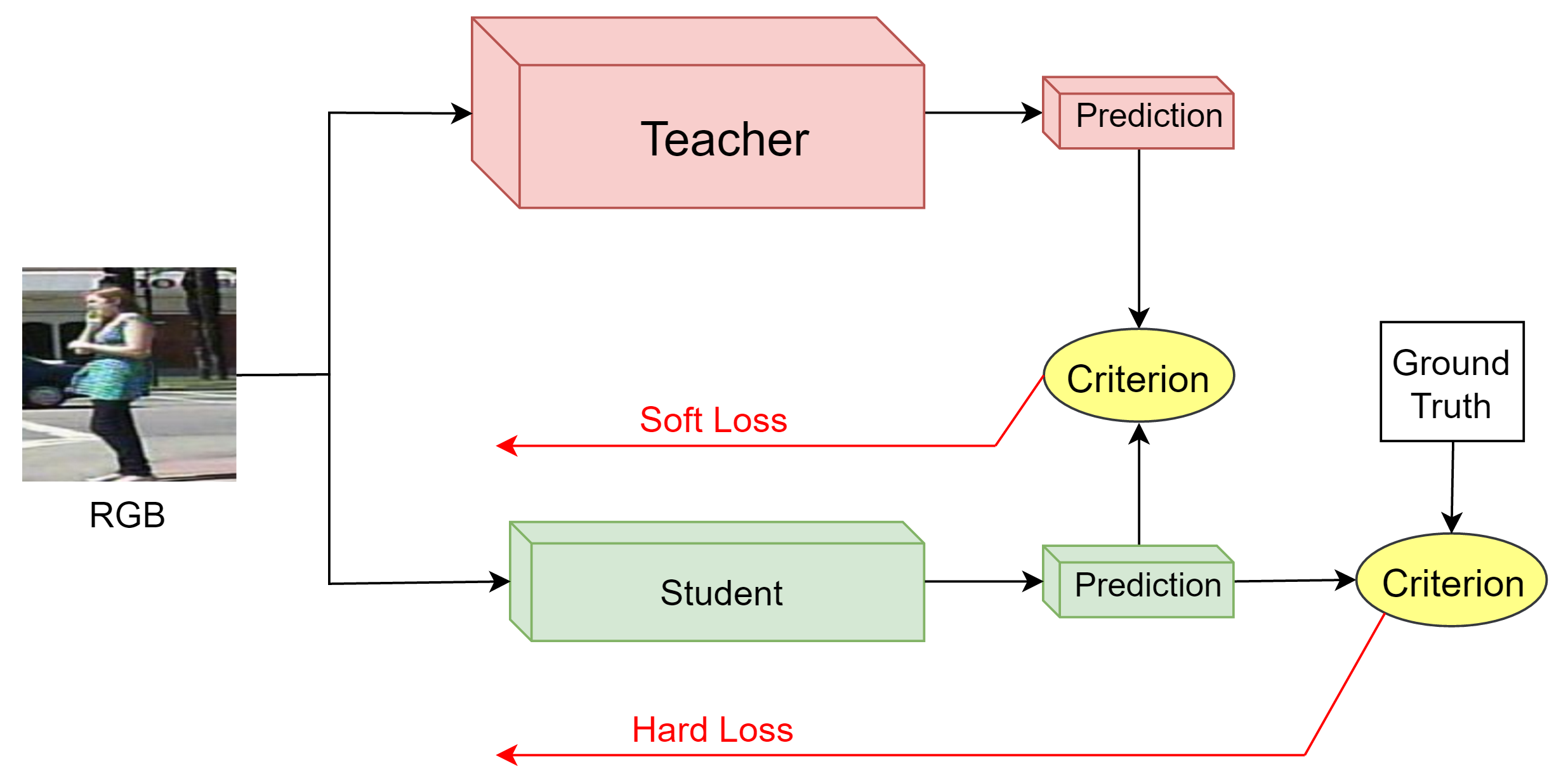}   
        \caption{Standard: student network learns from teacher guidance~(soft loss) and ground truth~(hard loss).}
        \label{figure:kd}
    \end{subfigure}
    \begin{subfigure}[b]{\linewidth}  
        \centering 
        \includegraphics[width=\linewidth]{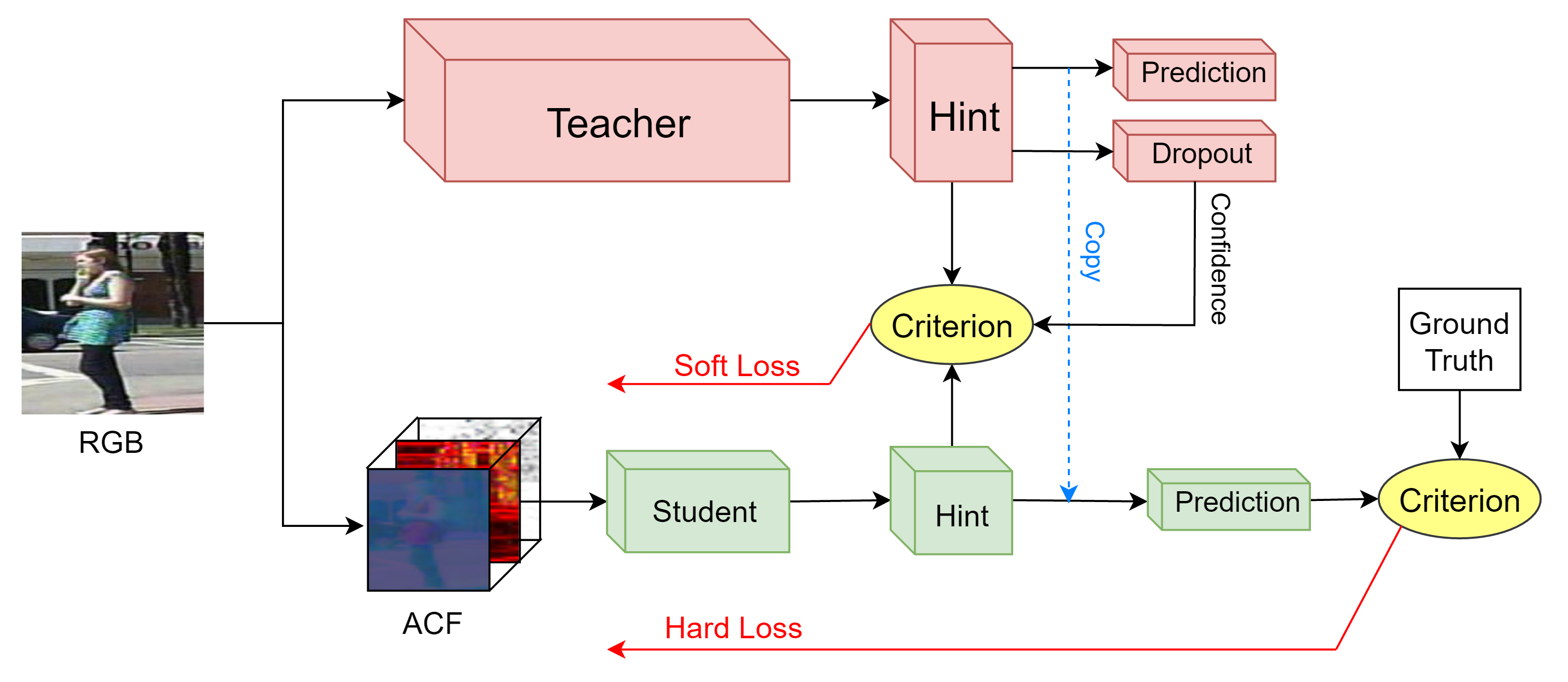}   
        \caption{Ours: student network uses ACF features as input and learns from teacher's hint layer outputs and covariances.}
        \label{figure:pipeline}
    \end{subfigure}
    \caption{Comparison between standard Knowledge Distillation~(a) and our pipeline~(b).}
\end{figure}

\section{Introduction}

State-of-the-art deep convolutional neural networks are extremely large and require a vast amount of resources. For example, the classic VGG-16 image classification network~\cite{simonyan2014very} contains 138~million parameters, and the more recent ResNet-200~\cite{he2015deep} still contains over 60~million parameters. 

This holds true not just for image classification. For instance, at the time of writing, the top three approaches for pedestrian detection as measured on the Caltech Pedestrian Dataset~\cite{Dollar2012PAMI} consist of MSCNN~\cite{cai16mscnn}, RPN+BF~\cite{zhang2016faster}, both built upon the Faster-RCNN~\cite{renNIPS15fasterrcnn} architecture containing over 100~million parameters, and SA-FastRCNN~\cite{li2015scale} which features a network with over 30~million parameters. The larger a network is, the more disk space, memory, and energy it consumes, and the slower it is to use.

These large networks contain many redundant parameters~\cite{Cun90optimalbrain,NIPS2013_5025}, so in theory they could be much smaller. To demonstrate this, we adopt Knowledge Distillation~(KD)~\cite{hinton2015distilling} to train a small \emph{student} network to mimic the large \emph{teacher} network. 


KD was developed for classification on the ImageNet dataset~\cite{russakovsky2015imagenet} with the idea that the 1000-dimensional prediction from the teacher is much more informative than the single ground truth label. But for pedestrian classification where there are only 2 outputs (pedestrian / no pedestrian), this difference is much less pronounced. To increase the dimensionality of the data that the student tries to learn, we propose introducing just before the final layer a reasonably sized fully-connected hint layer whose outputs we try to match.

The teacher network does not always predict the correct label. The policy behind KD is to train the student to mimic the teacher regardless of the mistakes. However, if the teacher has an estimate of how confident its prediction is, then the student could make more informed decisions. Intuitively, if the teacher reports that it is very confident about its prediction, then the student should trust the teacher more. To produce a measure of teacher confidence, we follow the insights in \cite{DBLP:journals/corr/GalG15} and utilize dropout at test time to fit a Gaussian distribution to the teacher outputs. We then propose a loss function that incorporates this information when learning from the teacher.

Hand-designed features are very popular in traditional computer vision. However, these features have not seen much success when used with deep learning~\cite{DBLP:journals/corr/HosangOBS15}. One theory is that deep networks are able to learn features internally that outperform the hand-designed features. If this is indeed the case, we hypothesize that small networks may not have the capacity to do so, and traditional feature extraction may improve small networks. We investigate this using Aggregate Channel Features~(ACF)~\cite{dollar2014fast} which are popular and proven features for pedestrian detection.

\subsection{Contributions}
In this paper we propose to use Knowledge Distillation to compress a large network for pedestrian classification. We explore variations on the training process by learning from the outputs of a hint layer inserted before the final fully-connected layer, introducing a loss function that takes into account output covariances, and using Aggregate Channel Features as input.

We show the effect of these modifications on student networks of various sizes, and show that they outperform both training from scratch and standard KD. We produce a model with only 157K parameters that outperforms AlexNet~\cite{NIPS2012_4824} which has over 57M parameters, 360$\times$ as many.

\section{Related Works}

\noindent\textbf{Network Pruning}\: The earliest studies into network size reduction came in the form of weight pruning, motivated by the need for regularization. These methods use the magnitude of the weights~\cite{hanson1989comparing} or the Hessian of the loss function~\cite{Cun90optimalbrain, hassibi1993optimal} to prune away less useful weights. Apart from pruning weights, Srinivas and Babu~\cite{DBLP:conf/bmvc/SrinivasB15} devised a method for pruning neurons directly without the use of any training data. These pruning approaches remove a significant amount of the uninformative parts of the network and results in lower computation costs and storage requirements.

\noindent\textbf{Parameter Sharing}\: Han~\etal~\cite{han2015deep} introduced a multi-step pipeline with pruning, weight clustering and Huffman encoding. An orthogonal approach uses hashing or bucketing to quantize various parts of the model~\cite{chen2015compressing, moons2016energy}. Cheng~\etal~\cite{DBLP:conf/iccv/ChengYFKCC15} enforce a circulant matrix model on the fully-connected layers to exploit faster computation and smaller model size via Fast Fourier Transforms. By quantizing and sharing parameters, the amount of space needed to store the network representation is reduced.

\noindent\textbf{Matrix Decomposition}\: Neural network weights can be treated as matrices and compressed through matrix decomposition. Denil~\etal~\cite{NIPS2013_5025} use a low rank decomposition of the weight matrices together with a sparse dictionary learned from an autoencoder to reduce the number of parameters. Novikov~\etal~\cite{NIPS2015_5787} apply the Tensor-Train decomposition~\cite{Oseledets:2011:TD:2079141.2079149} to compress the weight matrices in the fully-connected layers.

\noindent\textbf{Transfer Learning}. While the above methods compress an existing network directly, the underlying architecture remains bulky with the same wide and depth as before. An alternative is to consider transferring the knowledge to a new smaller network. This produces a much more compact model with dense weights instead of sparse weights. Moreover, it is possible to then apply the above methods on top of the new network.

Ba and Caruana~\cite{ba2014deep} showed that it is possible to train a shallower but wider student network to mimic a teacher network, performing almost as well as the teacher. Hinton~\etal~\cite{hinton2015distilling} generalized this idea by training the student to learn from both the teacher and from the training data, naming this process Knowledge Distillation~(KD). They demonstrated that students trained this way outperform those trained directly using only the training data. FitNets~\cite{romero2014fitnets} use Knowledge Distillation with intermediate hint layers to train a thinner but deeper student network containing fewer parameters that outperforms even the teacher network. However, to the best of our knowledge, this approach has yet to be applied to training a network that is both thinner and shallower.

\section{Knowledge Distillation}

The process of Knowledge Distillation~(KD) for classification networks is to train the student  from the predictions of the teacher network in addition to the ground truth \emph{hard targets} (Figure~\ref{figure:kd}). However, with a standard soft maximum~(softmax) classification layer, the teacher predictions will often be very similar to the hard targets with one class having probability close to 1 and the other classes having probabilities close to 0. So, instead, a variant of the softmax function which includes a temperature parameter $T$ is used instead to produce \emph{soft targets}.

\begin{equation}
\text{softmax}\left(\mathbf{L}, T\right) = \frac{\exp\left(\mathbf{L}/T\right)}{\sum_j \exp\left(L_j/T\right)}
\end{equation}

When $T=1$, this is the standard softmax function, while higher values of $T$ produce a  smoother probability distribution over the classes. $\mathbf{L}$ are the input logits to the softmax layer, and are also the outputs of the fully-connected layer before it.

The loss function $\mathcal{L}$ used for training the student is a combination of the soft loss $\mathcal{L}_\text{soft}$, the cross-entropy loss between the soft outputs of the student and teacher, as well as the hard loss $\mathcal{L}_\text{hard}$, the standard classification cross-entropy loss between the student outputs and the ground truth labels.

\begin{align}
\mathcal{L}_\text{soft} &= \mathcal{H}\left(\text{softmax}(\mathbf{L}_S,T),\text{softmax}(\mathbf{L}_T,T)\right) \label{softloss}\\
\mathcal{L}_\text{hard} &= \mathcal{H}\left(\mathbf{Y}_S, \mathbf{Y}_\text{GT}\right) = \mathcal{H}\left(\text{softmax}(\mathbf{L}_S,1),\mathbf{Y}_\text{GT}\right)\\
\mathcal{L} &= \mathcal{L}_\text{soft} + \lambda\mathcal{L}_\text{hard}
\end{align}

\section{Augmenting Knowledge Distillation}

A graphical outline of our pipeline can be found in Figure~\ref{figure:pipeline}. Here we explain the various parts of the pipeline and the motivations behind them.

\subsection{Hint Layer}

KD was developed for ImageNet classification with the idea that the 1000-dimensional prediction from the teacher is much more informative than the single ground truth label. But for pedestrian classification where there are only 2 outputs (pedestrian / no pedestrian), this difference is much less pronounced. Since the output of a softmax function sums up to 1 for every value of $T$, the soft targets are actually only 1-dimensional.

To increase the dimensionality of the data that the student learns from, we introduce a hint layer, a fully-connected~(FC) layer with 64 outputs in front of the final FC layer, and train the student to match the outputs of the hint layer instead. If the student network can perfectly match the hint layer outputs, then just by copying over the teacher's final FC layer, the student will be able to mimic the teacher's outputs. Even if the student cannot perfectly match the hint layer outputs, the weights from the teacher's final FC layer still serve as a good initialization for the student's final FC layer, which will be fine-tuned through the hard loss coming from the ground truth labels. In this work, we assume that the hint layer is the same size for both the teacher and student so that interpolation is not required.

This idea of matching earlier hint layers has been explored in FitNets~\cite{romero2014fitnets}. However, they choose to match a layer in the middle of the model to provide additional guidance in training a thinner but deeper student network. We instead propose this idea in order to increase the amount of information obtained from the teacher model in cases where there are only a small number of output classes.

The outputs of this hint layer cannot be interpreted as a probability distribution, so cross-entropy loss is not applicable. Instead, we use mean squared error loss as the soft loss.

Care must be taken when the activation for the hint layer is a rectified linear~(ReLU) nonlinearity, in which case it is advised to match the values before passing them through the ReLU function. This is because the ReLU function discards information of negative values, and also because the gradient for where the student predicts a negative value is ignored, leading to instabilities in training.

\subsection{Learning With Confidence}

There will be cases where the teacher makes mistakes and predicts differently from the ground truth. The policy behind KD is to train the student to mimic the teacher regardless of the mistakes, relying on the hard losses to nudge the outputs towards the correct label. This results in a tension between the soft and hard losses, each producing a gradient for the opposite label.

This tension can be relaxed slightly if the teacher has an estimate of how confident its prediction is. Intuitively, if the teacher reports that it is very confident about its prediction, then the student should trust the teacher more, and if the teacher instead reports that it is not confident about its prediction, then the student should balance mimicking the teacher with predicting the correct label. The underlying assumption is that the teacher is more likely to be confident about examples that they predict correctly. There will be cases where the teacher is very confident yet mistaken, but we believe that it is important for the student not to disregard the teacher in these cases.

In \cite{DBLP:journals/corr/GalG15}, the authors draw a theoretical link casting dropout as a Bayesian approximation of Gaussian Processes. Following their ideas, we enable dropout during test time and forward the same input through the model $N$ times. Each pass can be thought of as the output of a single model sampled from an ensemble. From this, the sample mean $\bar{\mathbf{Y}}$ and covariance $\hat{\mathbf{\Sigma}}$ of the outputs of the ensemble can be estimated. 

By doing so, we are fitting a multivariate Gaussian distribution to the teacher outputs, from which it is possible to measure the likelihood of the student output as being drawn from the distribution. In particular, the likelihood of the student output is:

\begin{equation}
p(\mathbf{Y}_S) = \frac{\exp\left(-\frac{1}{2}\left(\mathbf{Y}_S-\bar{\mathbf{Y}}_T\right)^T\mathbf{\hat{\Sigma}}^{-1}\left(\mathbf{Y}_S-\bar{\mathbf{Y}}_T\right)\right)}{\sqrt{(2\pi)^k|\mathbf{\hat{\Sigma}}|}}
\label{eqn:multgauss}
\end{equation}

Maximizing the log-likelihood of Equation~\ref{eqn:multgauss} is equivalent to minimizing the following loss function:

\begin{equation}
\mathcal{L}_\text{soft} = \left(\mathbf{Y}_S-\bar{\mathbf{Y}}_T\right)^T\mathbf{\hat{\Sigma}}^{-1}\left(\mathbf{Y}_S-\bar{\mathbf{Y}}_T\right)
\label{eqn:soft_loss}
\end{equation}

This function is the square of the Mahalanobis distance. Compared to the mean-square distance, it is smaller along dimensions of high variability, consistent with our idea of reporting smaller gradients for outputs that the teacher is not confident in.

One limitation of our method lies in the dimension of the covariance matrix. Since the loss function requires the inversion of covariance matrix, the number of samples $N$ must be larger than the dimensionality of the teacher's output. However, the output from the time consuming convolutional layers can be cached, and only the last few layers with dropout need multiple passes, so the additional overhead during training is low.

\subsection{Hand-designed Features as Input}

Before deep learning became mainstream, computer vision was dominated by the use of specialized features for each task discovered through extensive experimentation. For example, the advent of HOG features~\cite{dalal2005histograms} was ground-breaking in the development of pedestrian detection, and the introduction of Integral Channel Features~\cite{dollar2009integral} brought about another revolution, leading to the discovery of many derivative features such as Aggregate Channel Features~\cite{dollar2014fast} and Checkerboards features~\cite{zhang2015filtered}, the latter of which is competitive with state-of-the-art.

These hand-designed features are largely ignored in deep learning. In \cite{DBLP:journals/corr/HosangOBS15}, the authors found that there was no improvement in neural networks trained using hand-designed features compared to those trained using raw RGB images as input. However, their model was large, so it is possible that it was able to learn features internally that outperform the hand-designed features. The same might not be true for a small model, in which case it may be reasonable to expect that by using these hand-designed features as input, the small model can be improved. The use of hand-designed features as input can also be thought of as attaching a fixed layer to the front of the network, pre-trained through years of human research. 

For this reason, we explore training our student networks using Aggregate Channel Features~(ACF) as input. We choose ACF because it offers a good trade-off between detection accuracy and speed, taking less than 10ms to compute for a $640~\times~480$~image on a single CPU~\cite{dollar2014fast}. 

ACF consist of 10 channels: the LUV color channels, gradient magnitude, and six oriented gradient bins. The input image is first converted into these 10 channels, then, within each channel, pixels are divided into 4x4 blocks and summed.

Note that when we train the student using ACF features as input, the input to the teacher remains the original RGB image. Whether the student is trained on RGB or ACF, they learn from the exact same teacher.

\section{Experimental setup}

\subsection{Dataset}
We perform all training and evaluation on the Caltech Pedestrian Dataset~\cite{Dollar2012PAMI}. 
Following standard practice, we use the first 5 sequences as the training set, the 6th sequence as the validation set, and the last 5 sequences as the test set. 

We follow the setup of Caltech10x in \cite{DBLP:journals/corr/HosangOBS15} and sample every 3rd frame for training. We use the Reasonable configuration when testing on the Caltech test set, which samples every 30th frame and includes only pedestrians without significant occlusion with a minimum height of 50 pixels and excludes the labels ``people'' and ``person?''. Evaluation is performed using the official evaluation script, which computes a curve of the logarithm of the number of false positives per image versus the miss rate. A value for the log-average miss rate is also calculated, and a lower value indicates a better result. 

Our training set uses ground truth patches as well as patches with Intersection-over-Union~(IoU) greater than 0.5 as positive patches, and patches with IoU less than 0.5 as negative patches. There are 31,129 positive patches and 748,139 negative patches in the training set.

\subsection{Region Proposal}
We follow \cite{DBLP:journals/corr/HosangOBS15} and use the publicly available SquaresChnFtrs~\cite{Benenson2014Eccvw} region proposals. Using the same region proposal as \cite{DBLP:journals/corr/HosangOBS15} also gives us a fair comparison between AlexNet and our student network. The oracle miss rate of this region proposal is 13.2\% at 2.43 false positives per image.

\subsection{Models}

\noindent\textbf{Teacher Network}\: For our teacher network, We use pre-activation ResNet-200~\cite{he2016identity} pre-trained on ImageNet, augmented with dropout and a 64-dimensional hint layer, then fine-tuned on our training set.

\noindent\textbf{Student Network}\: We use pre-activation ResNet-18~\cite{he2016identity} pre-trained on ImageNet augmented with a 64-dimensional hint layer as the basis for our student networks. We experiment with three versions: unmodified ResNet-18, ResNet-18-Thin which cuts the number of channels for every layer in half, and ResNet-18-Small which fixes every layer at 32 channels. The compression rates of these models can be found in Table~\ref{table:compression}.

\begin{table}[t]
  \centering
  \begin{tabular}{|l|c|c|} \hline
             & \textbf{ResNet-200}   & \textbf{ResNet-18}                    \\ \hline
    \multirow{2}{*}{conv1}  
             & $7\times7\times64$, stride 2 & $7\times7\times64$, stride 2   \\
             & $3\times3$ pool, stride 2    & $3\times3$ pool, stride 2      \\ \hline
    conv2\_x &  $\begin{bmatrix}
                    3\times3\times64 \\
                    3\times3\times64 \\
                    3\times3\times256 \\
                \end{bmatrix}$ x3  
                
             &  $\begin{bmatrix}
                    3\times3\times64 \\
                    3\times3\times64 \\
                \end{bmatrix}$ x2  \\
                 \hline
    conv3\_x &  $\begin{bmatrix}
                    3\times3\times128 \\
                    3\times3\times128 \\
                    3\times3\times512 \\
                \end{bmatrix}$ x24   
                
             &  $\begin{bmatrix}
                    3\times3\times128 \\
                    3\times3\times128 \\
                \end{bmatrix}$ x2  \\
                 \hline
    conv4\_x &  $\begin{bmatrix}
                    3\times3\times256 \\
                    3\times3\times256 \\
                    3\times3\times1024 \\
                \end{bmatrix}$ x36    
                
             &  $\begin{bmatrix}
                    3\times3\times256 \\
                    3\times3\times256 \\
                \end{bmatrix}$ x2  \\
             \hline
    conv5\_x & $\begin{bmatrix}
                    3\times3\times512 \\
                    3\times3\times512 \\
                    3\times3\times2048 \\
                \end{bmatrix}$ x3
                
             &  $\begin{bmatrix}
                    3\times3\times512 \\
                    3\times3\times512 \\
                \end{bmatrix}$ x2  \\
              \hline
    \multirow{4}{*}{classifier}
             & avgpool & avgpool \\
             & dropout & FC(512, 64, ReLU) \\
             & FC(2048, 64, ReLU) & FC(64, 2, softmax) \\
             & FC(64, 2, softmax) & \\ \hline
   \end{tabular}
   \captionsetup{width=\linewidth}
   \caption{Teacher and Student model architectures for $224~\times~224$ input patches. Pool refers to a max-pooling layer, FC refers to a fully-connected layer, and avgpool refers to a global average pooling layer. All convolutional layers include batch normalization and a ReLU activation. The first convolution layer for conv\{3,4,5\} have a stride of 2.}
   \label{table:models}
\end{table}

\begin{table}[ht]
  \captionsetup{width=\linewidth}
  \centering
  \begin{tabular}{|l|c|c|} \hline
   \textbf{Model}       & \textbf{\#Parameters}   & \textbf{Compression}   \\ \hline
   ResNet-200           & 63M                     & $1\times$              \\ \hline
   ResNet-18            & 11M                     & $6\times$              \\ \hline
   ResNet-18-Thin       & 2.8M                    & $22\times$             \\ \hline
   ResNet-18-Small      & 157K                    & $400\times$            \\ \hline
   AlexNet              & 57M                     & $1\times$              \\ \hline
   \end{tabular}
   \caption{Comparison of the sizes of our various models and AlexNet.}
    \label{table:compression}
\end{table}

\subsection{Training configuration}

Training is performed through stochastic gradient descent with Nesterov Momentum 0.9 and weight decay 0.0005. We use a batch size of 16, an epoch size of 1000 iterations, a learning rate of 0.01 dropping by a factor of 5 every 20 epochs, and a total of 70 epochs. Since there are many more negatives patches than positive patches in our training set, we force a positive to negative ratio of 1:3 for each training batch.

The inputs to the teacher network are 224x224x3 RGB patches, and the inputs to the student networks are either 224x224x3 RGB patches or 224x224x10 ACF patches. Patches are scaled by warping them to fit the input size, and RGB inputs are normalized using ImageNet mean and standard deviation. During training, patches are randomly flipped horizontally. The extraction of ACF features occur after the flip.

When combining soft and hard losses, hard losses are weighted with a lambda of 0.5. For dropout during testing, we use a probability of 0.5. When estimating the covariance of teacher output, we forward each input 200 times. 

The models are trained with the Torch framework on a NVIDIA Titan X GPU with 12GB memory.

\section{Experimental Evaluation}

\begin{figure}[t]
\centering
\includegraphics[width=\linewidth]{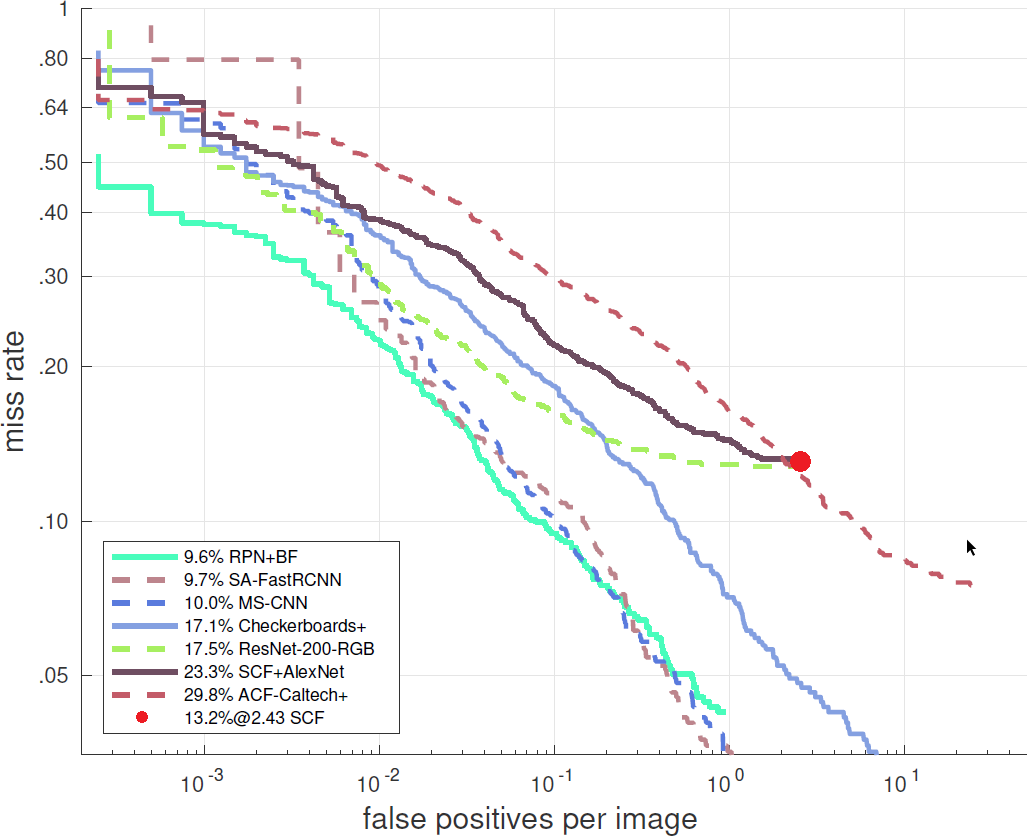}
\caption{Comparison of our teacher model and region proposal with state-of-the-art and select models.}
\label{fig:state-of-the-art}
\end{figure}

All of the evaluation results are reported on the Reasonable subset of the Caltech test set using the model at the epoch with the lowest log-average miss rate on the Reasonable subset of the Caltech validation set.

For all of the student models tested, the best performing configuration was to add a hint layer, use teacher confidence, and train with RGB inputs. A summary of our results can be found in Table~\ref{table:summary}. In the following sections, we break down the contribution from each of our innovations.

\subsection{Baselines}

\begin{table}[h]
  \captionsetup{width=\linewidth}
  \centering
  \begin{tabular}{|l|c|c|} \hline
   \textbf{Model}       & \textbf{Direct}         & \textbf{KD}            \\ \hline
   ResNet-200           & 17.5\%                  & ---                    \\ \hline
   ResNet-18            & 19.1\%                  & \textbf{18.6}\%                 \\ \hline
   ResNet-18-Thin       & \textbf{22.0}\%                  & 22.8\%                 \\ \hline
   ResNet-18-Small      & \textbf{24.5}\%                  & 24.8\%                 \\ \hline
   AlexNet              & 23.3\%                  & ---                    \\ \hline
  \end{tabular}
  \caption{Comparison of log-average miss rate when trained directly from the ground truth labels versus when trained through standard Knowledge Distillation with $T=2$. Results for AlexNet are from \cite{DBLP:journals/corr/HosangOBS15}.}
  \label{table:baselines}
\end{table}

Table~\ref{table:baselines} compares the various models trained directly from ground truth as well as the student models trained using standard Knowledge Distillation on the softmax logits. The temperature used for KD, $T=2$, was picked after testing multiple values.

\begin{figure}[t]
\centering
\includegraphics[width=\linewidth]{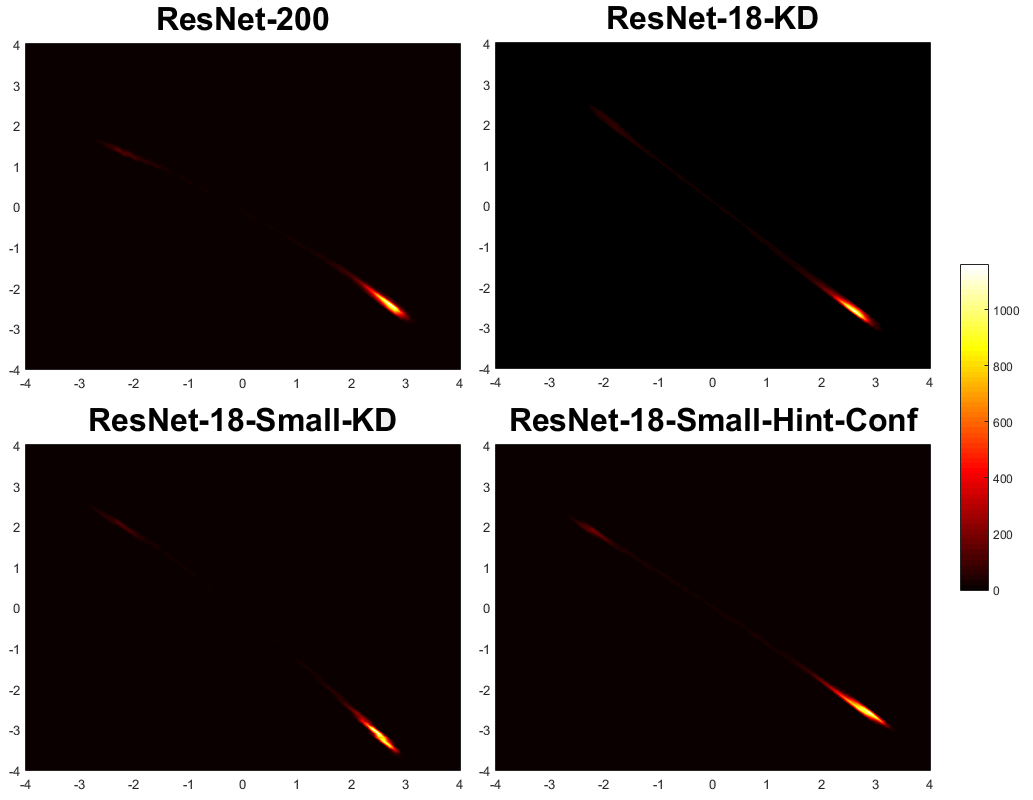}
\caption{Histogram of the two output logits on our test set for various models. The horizontal axis is the logit corresponding to a ``no pedestrian'' prediction, and the vertical axis to a ``pedestrian'' prediction.}
\label{fig:dist}
\end{figure}

Standard KD does not seem to work for the smaller models. In Figure~\ref{fig:dist}, we visualize the histogram of the distribution of the two output logits on the entire test set for the teacher~(ResNet-200), large~(ResNet-18) and small~(ResNet-18-Small) students trained via standard KD, and the small student trained with our full pipeline. Only the small student trained via standard KD is visibly different.

This difference highlights an important property of standard KD. With or without a temperature, the softmax function normalizes its outputs to sum to 1, and two different inputs can result in the same output. This is not necessarily a problem, but we hypothesize that with very small student models, for problems with very few output labels, standard KD does not offer enough guidance to be superior to training from ground truth labels.

\subsection{Hint Layer}

\begin{table}[h]
  \captionsetup{width=\linewidth}
  \centering
  \begin{tabular}{|l|c|c|} \hline
   \textbf{Model}       & \textbf{Direct}         & \textbf{Hint} \\ \hline
   ResNet-200           & 17.5\%                  & ---                      \\ \hline
   ResNet-18            & 19.1\%                  & \textbf{18.1}\%                   \\ \hline
   ResNet-18-Thin       & 22.0\%                  & \textbf{20.4}\%                   \\ \hline
   ResNet-18-Small      & 24.5\%                  & \textbf{23.1}\%                   \\ \hline
  \end{tabular}
  \caption{Comparison of log-average miss rate when trained directly from the ground truth labels versus when matching hint layer outputs.}
  \label{table:fcmatch}
\end{table}

As reported in Table~\ref{table:fcmatch}, adding a hint layer improves training for all student models. This enforces the idea that increasing the amount of information used for training models is beneficial.

It would be interesting to explore the effect of varying the size of the hint layer, but that is unfortunately outside the scope of this work. Is there a point where the hint layer is too large and dominated by noise instead of useful data?

\subsection{Learning With Confidence}

\begin{table}[h]
  \captionsetup{width=\linewidth}
  \centering
  \begin{tabular}{|l|c|c|c|} \hline
   \textbf{Model}   & \textbf{Direct}  & \textbf{Conf} & \shortstack{\textbf{Hint + Conf}\\\textbf{(Ours)}} \\ \hline
   ResNet-200       & 17.5\%           & ---              & ---                       \\ \hline
   ResNet-18        & 19.1\%           & 18.2\%           & \textbf{18.0}\%                    \\ \hline
   ResNet-18-Thin   & 22.0\%           & 20.7\%           & \textbf{20.3}\%                    \\ \hline
   ResNet-18-Small  & 24.5\%           & 23.7\%           & \textbf{22.4}\%                    \\ \hline
  \end{tabular}
  \caption{Comparison of log-average miss rate when trained directly from the ground truth labels versus when trained with teacher output covariances, estimated either from the softmax logits or from the hint layer outputs.}
  \label{table:dropout}
\end{table}

Table~\ref{table:dropout} shows that estimating output covariances and training with our proposed loss function improves the student models. There is slightly more improvement if the covariances are estimated from the 64-dimensional hint layer outputs instead of the 2-dimensional softmax logits.

This shows that there is indeed more information in the teacher network that can be extracted when treated as an ensemble using dropout.

\subsection{Hand-designed Features as Input}

\begin{table}[h]
  \captionsetup{width=\linewidth}
  \centering
  \setlength\tabcolsep{3pt}
  \begin{tabular}{|l|c|c|c|} \hline
   \textbf{Model}   & \textbf{RGB}  & \textbf{ACF} & \textbf{ACF + Hint + Conf} \\ \hline
   ResNet-200       & 17.5\%           & 19.6\%           & ---                       \\ \hline
   ResNet-18        & 19.1\%           & 21.4\%           & \textbf{18.7}\%                    \\ \hline
   ResNet-18-Thin   & 22.0\%           & 22.4\%           & \textbf{20.4}\%                    \\ \hline
   ResNet-18-Small  & 24.5\%           & 25.2\%           & \textbf{23.4}\%                    \\ \hline
  \end{tabular}
  \caption{Comparison of log-average miss rate when trained directly from the ground truth labels with RGB inputs, ACF inputs, and when trained with ACF inputs with teacher output covariances estimated from hint layer outputs.}
  \label{table:acf}
\end{table}

Our results in Table~\ref{table:acf} are consistent with \cite{DBLP:journals/corr/HosangOBS15} in that a network trained directly using features like ACF as input is worse than if it were trained with raw RGB inputs. This holds true even as the network size is significantly reduced, though the drop in performance for smaller models is not as severe.

Similar to the results for our previous models taking RGB inputs, introducing a hint layer and factoring in teacher confidence offers similar amounts of improvement to the models taking ACF input. However, the models with ACF inputs still fall short compared to those that take RGB inputs, and we can only conclude that ACF is worse than RGB as inputs to pedestrian detection networks.

\section{Network Analysis}

In this section, we compare ResNet-200~(teacher) with ResNet-18-Small-RGB-Hint-Conf~(student).

\begin{figure}[t]
\centering
\includegraphics[width=\linewidth]{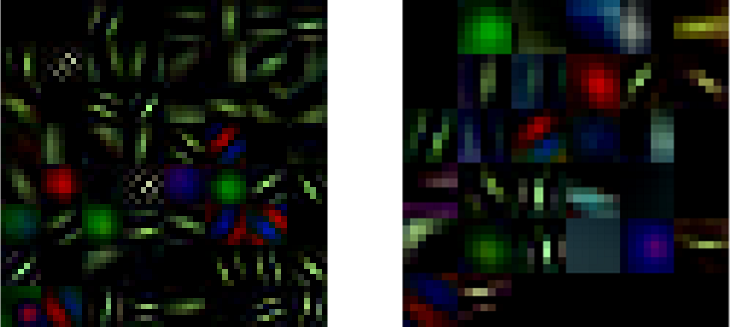}
\caption{Visualization of the first convolutional layer from teacher network~(left) and student network~(right). The contrast has been adjusted for visibility.}
\label{fig:conv}
\end{figure}

\noindent\textbf{Model Similarity}\: We visualize the weights of the first convolutional layer in Figure~\ref{fig:conv}. The first layer weights are slightly different, however, we can see similar shapes in the patterns. 




\begin{table}[h]
\centering
\begin{tabular}{c|c|c|}
\cline{2-3}
 & \begin{tabular}[c]{@{}c@{}}\textbf{Student}\\ \textbf{Correct}\end{tabular} & \begin{tabular}[c]{@{}c@{}}\textbf{Student}\\ \textbf{Fail}\end{tabular} \\ \hline
\multicolumn{1}{|c|}{\begin{tabular}[c]{@{}c@{}}\textbf{Teacher}\\ \textbf{Correct}\end{tabular}} & 73.74\% & 21.12\% \\ \hline
\multicolumn{1}{|c|}{\begin{tabular}[c]{@{}c@{}}\textbf{Teacher}\\ \textbf{Fail}\end{tabular}} & 0.60\% & 4.54\% \\ \hline
\end{tabular}
\caption{Number of correct/fail patches from our models.}
\label{table:failure}
\end{table}

\begin{figure*}[t]
    \centering
    \begin{subfigure}[b]{\textwidth}  
    \centering
        \includegraphics[width=0.15\textwidth]{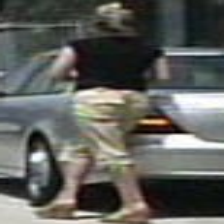}    
    \quad
        \includegraphics[width=0.15\textwidth]{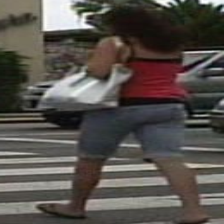}    
    \quad
        \includegraphics[width=0.15\textwidth]{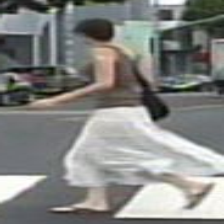}    
    \quad
        \includegraphics[width=0.15\textwidth]{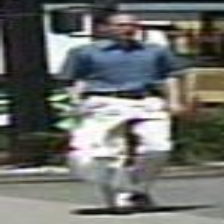}    
    \quad 
        \includegraphics[width=0.15\textwidth]{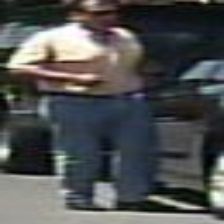}

    \caption{Positive patches: teacher predicts correctly but student predicts incorrectly.}
    \label{fig:student_fail1}
    \end{subfigure}

    \centering
    \begin{subfigure}[b]{\textwidth}  
    \centering
        \includegraphics[width=0.15\textwidth]{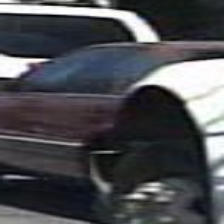}    
    \quad
        \includegraphics[width=0.15\textwidth]{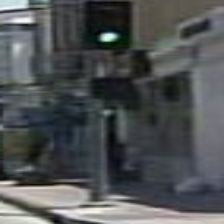}    
    \quad
        \includegraphics[width=0.15\textwidth]{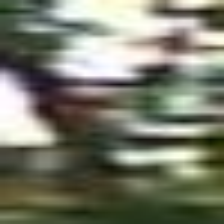}    
    \quad
        \includegraphics[width=0.15\textwidth]{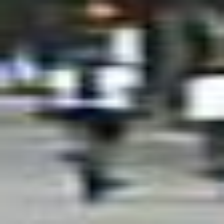}    
    \quad
        \includegraphics[width=0.15\textwidth]{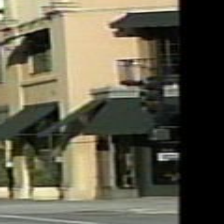}
    \caption{Negative patches: teacher predicts correctly but student predicts incorrectly.}
    \label{fig:student_fail2}
    \end{subfigure}

    \centering
    \begin{subfigure}[b]{\textwidth}  
    \centering
        \includegraphics[width=0.15\textwidth]{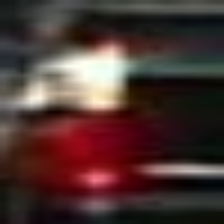}    
    \quad
        \includegraphics[width=0.15\textwidth]{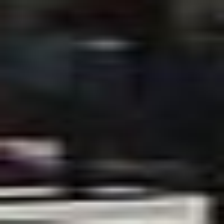}
    \quad
        \includegraphics[width=0.15\textwidth]{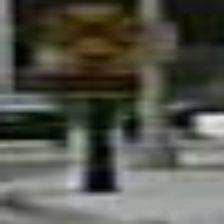}   
    \quad
        \includegraphics[width=0.15\textwidth]{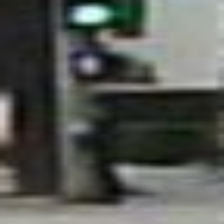}    
    \quad
        \includegraphics[width=0.15\textwidth]{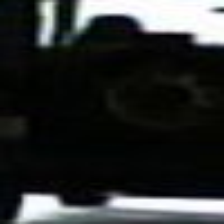}    
    \caption{Negative patches: student predicts correctly but teacher predicts incorrectly.}
    \label{fig:teacher_fail}
    \end{subfigure}
    
    \caption{Example of disagreements between teacher~(ResNet-200) and student~(ResNet-18-Small-RGB-Hint-Conf).}
\end{figure*}

\noindent\textbf{Failure Cases}\: We tabulate the correct and incorrect predictions for each model in Table~\ref{table:failure}. The teacher and student networks predict the same label 78.28\% of the time. We sample some example patches where the teacher predicts correctly but the student fails in Figure~\ref{fig:student_fail1},~\ref{fig:student_fail2} and the reverse in Figure~\ref{fig:teacher_fail}. The student did not predict any positive patches correctly that the teacher had predicted incorrectly. The patches where the student outperformed the teacher are indeed harder to classify, and could be a result of the student model being much smaller and thus more regularized.

\subsection{Resource Usage}

We report the resource usage during test time on a NVIDIA Titan X in Table~\ref{table:resource}. 

\begin{table}[h]
\centering
\begin{tabular}{|l|c|c|} \hline
\textbf{Model} & \textbf{Time} & \textbf{Memory} \\ \hline
ResNet-200 & 24ms & 5377MB \\ \hline
ResNet-18 & 3ms & 937MB \\ \hline
ResNet-18-Thin & 3ms & 633MB \\ \hline
ResNet-18-Small & 3ms & 565MB \\ \hline
Single Identity Layer & 0.02ms & 325MB \\ \hline
\end{tabular}
\caption{Prediction time and memory usage for a size 16 batch of 224x224 patches.}
\label{table:resource}
\end{table}

It appears that modern GPUs are not affected very much by the number of channels in convolutional layers, so while ResNet-18-Thin and ResNet-18-Small are much smaller in terms of the number of parameters, they are not significantly faster than ResNet-18. However, the memory usage is significantly decreased. Ignoring the fixed amount of memory used by the inputs and the system measured using a model with a single identity layer, ResNet-18-Small uses $(5377-325)/(565-325)=21\times$ less memory.

\begin{table*}[ht]
\centering
\begin{tabularx}{\textwidth}{|l|l|Y|Y|Y|Y|Y|Y|Y|Y|}
\hline
\multicolumn{2}{|c|}{\multirow{2}{*}{\textbf{Statistic}}} & \multicolumn{2}{c|}{\textbf{ResNet-200 (Teacher)}} & \multicolumn{2}{c|}{\textbf{ResNet-18}} & \multicolumn{2}{c|}{\textbf{ResNet-18-Thin}} & \multicolumn{2}{c|}{\textbf{ResNet-18-Small}} \\ \cline{3-10}
\multicolumn{2}{|c|}{} & \textbf{RGB} & \textbf{ACF} & \textbf{RGB} & \textbf{ACF} & \textbf{RGB} & \textbf{ACF} & \textbf{RGB} & \textbf{ACF} \\ \hhline{|==|=|=|=|=|=|=|=|=|}
\multirow{5}{*}{\rotatebox{90}{Log-Avg MR}}
& Direct & \textbf{17.5}\% & 19.6\% & 19.1\% & 21.4\% & 22.0\% & 22.4\% & 24.5\% & 25.2\% \\ \cline{2-10}
& KD & --- & --- & 18.6\% & --- & 22.8\% & 23.1\% & 24.8\% & --- \\ \cline{2-10}
& Conf & --- & --- & 18.2\% & --- & 20.7\% & --- & 23.7\% & --- \\ \cline{2-10}
& Hint & --- & --- & 18.1\% & --- & 20.4\% & 21.1\% & 23.1\% & --- \\ \cline{2-10}
& Hint+Conf & --- & --- & \textbf{18.0\%} & 18.7\% & \textbf{20.3\%} & 20.4\% & \textbf{22.4\%} & 23.4\% \\ \hhline{|==|==|==|==|==|}
\multicolumn{2}{|c|}{\#Parameters} & \multicolumn{2}{c|}{63M (1$\times$)} & \multicolumn{2}{c|}{11M (6$\times$)} & \multicolumn{2}{c|}{2.8M (22$\times$)} & \multicolumn{2}{c|}{157K (400$\times$)} \\ \hline
\multicolumn{2}{|c|}{Speed} & \multicolumn{2}{c|}{24ms (1$\times$)} & \multicolumn{2}{c|}{3ms (8$\times$)} & \multicolumn{2}{c|}{3ms (8$\times$)} & \multicolumn{2}{c|}{3ms (8$\times$)} \\ \hline
\multicolumn{2}{|c|}{Memory} & \multicolumn{2}{c|}{5052MB (1$\times$)} & \multicolumn{2}{c|}{612MB (8$\times$)} & \multicolumn{2}{c|}{308MB (16$\times$)} & \multicolumn{2}{c|}{240MB (21$\times$)} \\ \hline
\end{tabularx}
\caption{Summary of the results presented in this work. The configuration with the best log-average miss rate for each model is highlighted. Numbers in parenthesis indicate how much better the model is compared to the teacher.}
\label{table:summary}
\end{table*}

\section{Conclusion}

We have shown that there is indeed a lot of redundancy in large deep neural networks. We have shown that it is possible to train a student network that contains 400 times fewer parameters while only observing a drop in log-average miss rate of 4.9\%. The main gains of our approach utilizes the dimensionality of our new hint layers. We also described a method of obtaining a measure of confidence from the teacher network, and demonstrated that taking this information into account during training can lead to considerable gains. Our student models perform 8x faster than the teacher with 21x less memory usage.

{\small
\bibliographystyle{ieee}
\bibliography{main}
}

\end{document}